\documentclass[runningheads]{llncs}
\usepackage[T1]{fontenc}
\usepackage{graphicx}
\usepackage{cite}
\usepackage{amsmath,amssymb,amsfonts}
\usepackage{algorithmic}
\usepackage{graphicx}
\usepackage{textcomp}
\usepackage{xcolor}
\begin{document}
\title{Deep Learning CNN and Recurrence Analysis for Alpha–Gamma EEG Biomarkers in Fragile X Syndrome}
%
%
\author{Zag ElSayed\inst{1,2}\orcidID{0000-0001-9094-1469} \and
Payton Siekierski\inst{2}\orcidID{0009-0000-9681-5746} \and
Jack Yanchen Liu\inst{2}\orcidID{0009-0004-1340-0545} \and
Ernie Pedapati\inst{2}\orcidID{0000-0002-7954-5104}}
\authorrunning{Z. ElSayed et al.}
%
\institute{School of Information Technology (SoIT),
University of Cincinnati, Ohio, USA \and
Division of Child and Adolescent Psychiatry
Cincinnati Children’s Hospital Medical Center, Ohio, USA \\
\email{elsayezs@ucmail.uc.edu}\\
\email{siekiepe@mail.uc.edu}\\
\email{Yanchen.Liu@cchmc.org}\\
\email{Ernest.Pedapati@cchmc.org}
}
\maketitle              
\begin{abstract}
Fragile X Syndrome (FXS) is a neurodevelopmental disorder caused by reduced expression of fragile X mental retardation protein (FMRP), leading to disrupted synaptic plasticity, cortical hyperexcitability, and impaired network synchronization. Electroencephalography (EEG) provides a noninvasive window into these mechanisms and consistently reveals abnormalities in alpha (8--12 Hz) and gamma (30--100 Hz) oscillations that relate to inhibitory control, sensory processing, and cognition. This paper proposes a multi-representation deep learning framework for automated characterization of FXS EEG phenotypes by integrating convolutional neural networks (CNNs), long short-term memory (LSTM) networks, and recurrence plot (RP) analysis. Band-limited EEG signals are decomposed into alpha and gamma components and transformed into complementary representations, including temporal feature sequences, time-frequency maps, and RP images encoding the nonlinear recurrence structure. CNN modules learn discriminative spatial-spectral and dynamical textures from image-based representations, while LSTM modules model temporal modulation of oscillatory activity; a hybrid CNN-LSTM architecture jointly captures spatial, temporal, and nonlinear dependencies. Subject-independent evaluation demonstrates that the hybrid model outperforms single-modality baselines, with gamma features providing strong discriminative power and alpha-gamma integration yielding the best overall performance. These findings support deep learning with nonlinear representations as a scalable approach for EEG biomarker development in FXS, with potential utility for diagnosis, stratification, and treatment monitoring in translational settings.
\end{abstract}

\section{Introduction}
\begin{figure}[htbp]
\centerline{\includegraphics[width =0.95\linewidth]{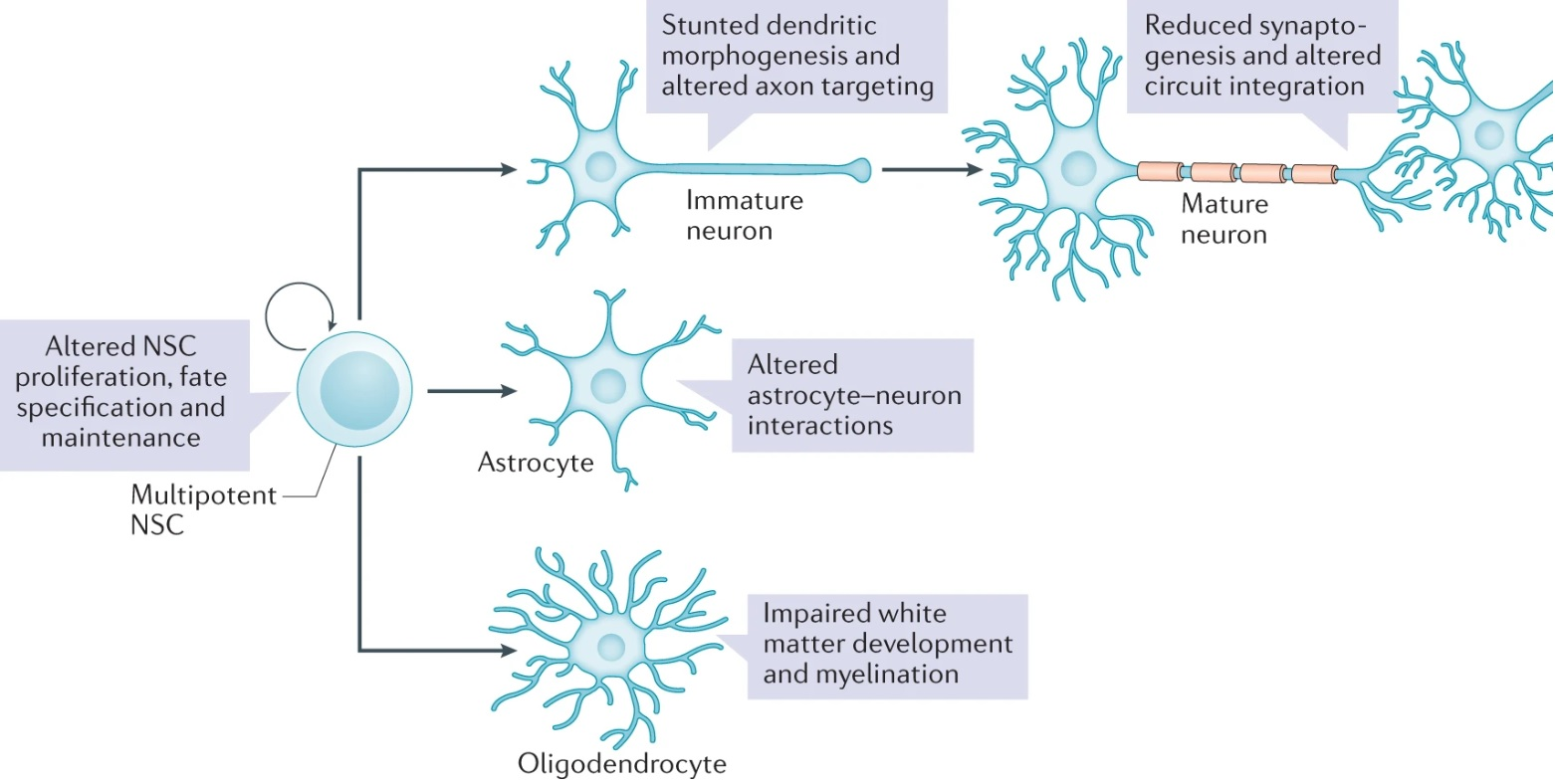}}
\caption{Fragile X syndrome is a neurodevelopmental disorder resulting from fragile X mental retardation protein (FMRP) deficiency, which adversely affects nearly every aspect of brain development and the functioning of many brain cell types \cite{Richter2021-uo}.}
\label{fig1}
\end{figure}

Fragile X Syndrome (FXS) is the most common inherited cause of intellectual disability and a leading monogenic contributor to autism spectrum disorders, resulting from a CGG trinucleotide repeat expansion in the \textit{FMR1} gene and subsequent deficiency of fragile X mental retardation protein (FMRP) \cite{Cornish2004-vr}. Reduced FMRP disrupts synaptic development, excitation-inhibition balance, and cortical network maturation, leading to characteristic cognitive, behavioral, and sensory impairments (Fig.~\ref{fig1}). Clinical manifestations are typically more severe in males, while females exhibit broader phenotypic variability.

Electroencephalography (EEG) provides a noninvasive window into the neurophysiological consequences of FMRP deficiency and has consistently revealed abnormal oscillatory activity in FXS, particularly within the alpha (8-12 Hz) and gamma (30-100 Hz) frequency bands. Alpha rhythms are associated with large-scale inhibitory control and network coordination, whereas gamma oscillations reflect local circuit excitability and information processing. Abnormalities in these bands are thought to reflect cortical hyperexcitability and impaired inhibitory regulation in FXS.

Despite robust group-level findings, extracting clinically meaningful EEG biomarkers in FXS remains challenging. EEG signals are nonstationary, noisy, and high-dimensional, with substantial inter-individual variability that limits the sensitivity of conventional linear spectral and connectivity analyses. Recent advances in machine learning (ML) and deep learning offer new opportunities to address these challenges by enabling automated feature learning and nonlinear modeling of complex neural dynamics.

In this work, we propose a unified deep learning framework that integrates convolutional neural networks (CNNs), long short-term memory (LSTM) networks, and recurrence plot (RP) representations to characterize alpha-gamma EEG abnormalities in FXS. By jointly modeling spatial-spectral structure, temporal dynamics, and nonlinear recurrence behavior, the proposed approach aims to improve discrimination between FXS and typically developing controls while providing a scalable foundation for EEG-based biomarker development.

\section{Background and Related Work}

\subsection{Neurophysiology of Fragile X Syndrome}
Fragile X Syndrome arises from reduced expression of FMRP, a protein critical for synaptic plasticity and activity-dependent regulation of excitatory and inhibitory circuits. Animal and human studies demonstrate that FMRP deficiency leads to cortical hyperexcitability, impaired inhibitory interneuron function, and altered network synchronization. These mechanisms manifest at the electrophysiological level as abnormal oscillatory dynamics and disrupted functional connectivity.

Additionally, it was observed that in the alpha, upper beta, gamma, and epsilon frequency bands, relative power differed by sex, whereas changes in theta and low beta power were similar in FXS males and females~\cite{Castren2003-uh}.
Due to these phenomena, several attempts at metadata analyses were proposed, such as the study in~\cite{Ethridge2017-si}, which synthesized data from multiple EEG studies on FXS, confirming that both alpha and gamma band abnormalities are consistently observed across different age groups and experimental conditions.

Additionally, it was observed that in the alpha, upper beta, gamma, and epsilon frequency bands, relative power differed by sex. In contrast, changes in theta and low beta power were similar in FXS males and females~\cite{Castren2003-uh}.
Due to these phenomena, several attempts at metadata analysis were proposed, such as the study in~\cite{Ethridge2017-si}, which synthesized data from multiple EEG studies on FXS and confirmed that both alpha and gamma band abnormalities are consistently observed across different age groups and experimental conditions.

\subsection{EEG Oscillatory Abnormalities in FXS}
\begin{figure}[htbp]
\centerline{\includegraphics[width =0.95\linewidth]{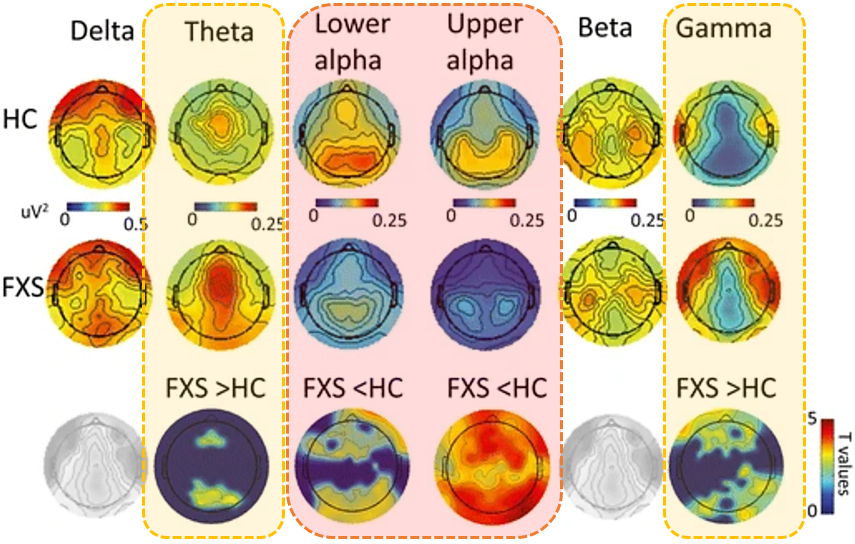}}
\caption{Permutation tests based on significant group differences in connection strength between Fragile X Syndrom (FXS) and healthy control (HC) individuals (p\textless 0.05) ~\cite{Wang2017}, reveal that FXS has more connectivity across electrodes in the gamma band but less within-band connectivity in the alpha (lower and upper) and delta range. \cite{Richter2021-uo, moskowitz2015uncovering}.}
\label{fig2}
\end{figure}

EEG studies in FXS have consistently reported alterations across multiple frequency bands, with the most robust findings observed in alpha and gamma activity. Increased resting-state gamma power, reduced gamma phase synchrony, and disrupted functional connectivity patterns have been linked to cortical hyperexcitability and impaired inhibitory control (Fig.~\ref{fig2}) \cite{Wang2017, Wilkinson2021-pf}. Developmental studies further suggest delayed maturation of alpha rhythms and persistent gamma abnormalities across the lifespan.

Comparative and meta-analytic studies confirm that alpha-gamma dysregulation is a reproducible hallmark of FXS  \cite{LIU2023100070} and distinguishes it from related neurodevelopmental conditions such as autism spectrum disorder, despite partial overlap in electrophysiological phenotypes \cite{van2012auditory, knyazev2012eeg, Liang2022-rn}.

\subsection{Computational and Machine Learning Approaches in FXS EEG}
Beyond descriptive spectral analyses, recent work has increasingly applied statistical modeling and machine learning to quantify FXS EEG phenotypes \cite{pedapati2025frontal}. Prior studies \cite{Kenny2022Biomarker} have explored guided feature selection, classification, and subgroup stratification using resting-state and task-evoked EEG measures \cite{Ethridge2019Auditory}, demonstrating the potential of computational approaches for diagnosis and biomarker discovery \cite{Ethridge2024, Knoth2018, Norris2025-vc}. Nonlinear complexity measures and connectivity modeling further indicate that FXS alters not only oscillatory power but also dynamical structure and stability \cite{Knoth2018}.

However, most existing approaches rely on handcrafted features and shallow classifiers, which may fail to capture higher-order spatial, temporal, and nonlinear dependencies present in EEG signals. Deep learning architectures, particularly CNNs for spatial and spectral learning, LSTMs for temporal modeling, and image-based representations such as recurrence plots, remain underexplored in the context of FXS.

\subsection{Motivation for the Present Work}
\begin{figure}[htbp]
\centerline{\includegraphics[width =0.9\linewidth]{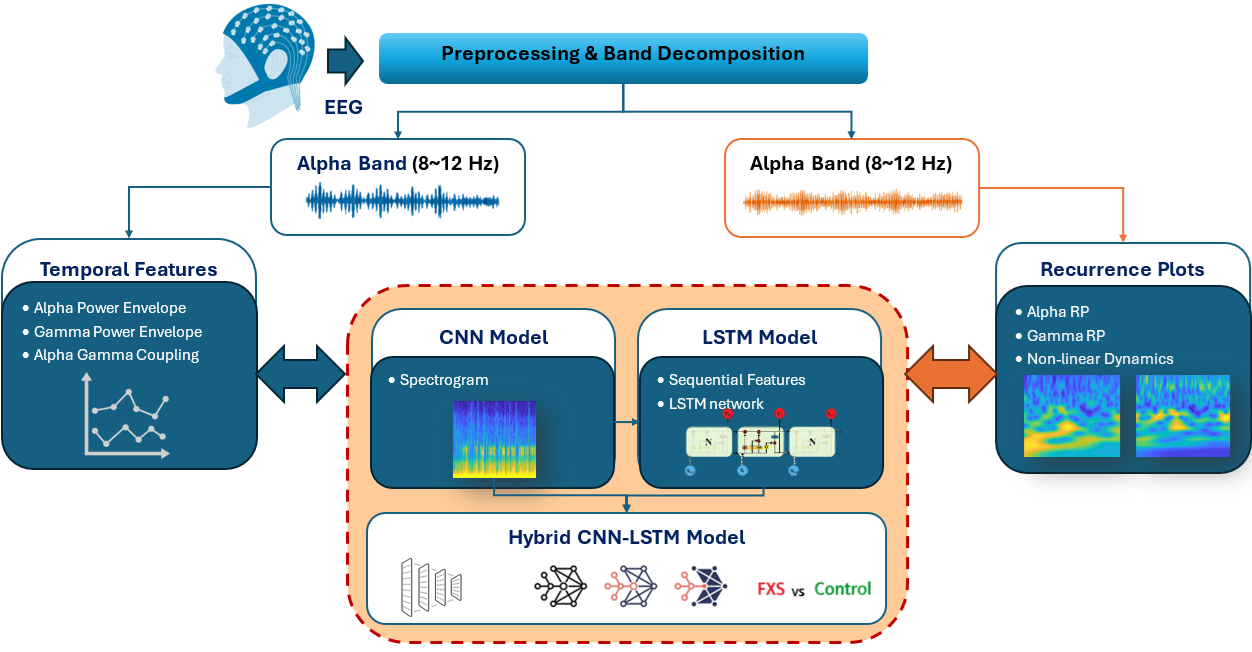}}
\caption{CNN–LSTM–RP pipeline for modeling alpha–gamma EEG abnormalities in Fragile X Syndrome.}
\label{fig3}
\end{figure}

Motivated by converging neurophysiological and computational evidence, this study introduces a deep learning framework that explicitly targets alpha-gamma dysregulation in FXS using complementary representations of EEG dynamics. By integrating CNN-based feature learning, LSTM-based temporal modeling, and recurrence plot analysis, shown in Fig.~\ref{fig3}, the proposed approach addresses key limitations of prior methods and advances toward automated, scalable EEG biomarkers for Fragile X Syndrome.

\section{Methodology}
Fig.~\ref{fig3} summarizes the proposed CNN-LSTM-recurrence plot framework. Let the multichannel EEG be
\begin{equation}
\mathbf{X} \in \mathbb{R}^{C \times T},
\label{eq1}
\end{equation}
where $C$ is the number of channels and $T$ the number of samples. After preprocessing $\mathcal{P}(\cdot)$, alpha and gamma components are extracted by band-pass filters:
\begin{equation}
\mathbf{X}^{\alpha}=\mathcal{B}_{\alpha}\bigl(\mathcal{P}(\mathbf{X})\bigr), \quad
\mathbf{X}^{\gamma}=\mathcal{B}_{\gamma}\bigl(\mathcal{P}(\mathbf{X})\bigr),
\label{eq2}
\end{equation}
where $\mathcal{B}_{\alpha}$ and $\mathcal{B}_{\gamma}$ denote alpha (8-12 Hz) and gamma (30-100 Hz) band-pass filters. Two representations are constructed: (i) temporal feature sequences and (ii) recurrence-plot images. Temporal features are
\begin{equation}
\mathbf{S}=\{s_t\}_{t=1}^{T'}, \quad s_t=f(\mathbf{X}^{\alpha}_t,\mathbf{X}^{\gamma}_t),
\label{eq3}
\end{equation}
where $f(\cdot)$ extracts bandpower and peak-frequency descriptors per window and $T'$ is the number of windows. Recurrence plots are
\begin{equation}
\mathbf{R}=\mathcal{R}(\mathbf{X}^{\alpha},\mathbf{X}^{\gamma}),
\label{eq4}
\end{equation}
where $\mathcal{R}(\cdot)$ denotes phase-space embedding and recurrence mapping. Feature learning is performed via
\begin{equation}
\mathbf{Z}_{\mathrm{CNN}}=\mathcal{F}_{\mathrm{CNN}}(\mathbf{R}), \quad
\mathbf{Z}_{\mathrm{LSTM}}=\mathcal{F}_{\mathrm{LSTM}}(\mathbf{S}),
\label{eq5}
\end{equation}
and the hybrid model is
\begin{equation}
\mathbf{Z}=\mathcal{F}_{\mathrm{LSTM}}\bigl(\mathcal{F}_{\mathrm{CNN}}(\mathbf{R})\bigr).
\label{eq6}
\end{equation}
Finally, classification is
\begin{equation}
\hat{y}=\sigma(\mathbf{W}\mathbf{Z}+\mathbf{b}),
\label{eq7}
\end{equation}
where $\mathbf{W}$ and $\mathbf{b}$ are trainable parameters and $\sigma(\cdot)$ is sigmoid/softmax.

\subsection{EEG Dataset and Experimental Paradigm}
EEG recordings were collected from individuals with Fragile X Syndrome (FXS) and age-matched typically developing (TD) controls using high-density montages ($C\geq64$) and sampling rates $\geq 500$ Hz. Data included resting-state (eyes-open/eyes-closed) segments and, when available, auditory paradigms. Only artifact-controlled segments exceeding a minimum duration threshold were retained to ensure stable estimation of alpha and gamma activity. Subject-independent partitioning was enforced such that all segments from the same subject appeared in exactly one split (train/validation/test), preventing data leakage and reflecting clinical deployment settings.

\subsection{EEG Preprocessing and Artifact Removal}
Signals were band-pass filtered (1-100 Hz, zero-phase FIR) with 60 Hz notch filtering. Noisy/flat channels were detected using variance/kurtosis criteria and interpolated as needed. Artifacts were removed using ICA by rejecting components associated with ocular, muscle, cardiac, and channel noise based on topography, temporal characteristics, and spectra. Cleaned signals were reconstructed, re-referenced to the common average, and z-scored per channel to reduce inter-session amplitude bias.

\subsection{Alpha and Gamma Band Decomposition and Feature Construction}
After preprocessing, band-limited signals were extracted using zero-phase band-pass filters for alpha (8-12 Hz) and gamma (30-100 Hz). Welch PSD estimates yielded bandpower, peak frequency, and peak power per epoch/channel. Temporal dynamics were captured using Hilbert envelopes segmented into fixed windows to form sequences for LSTM modeling. Cross-band descriptors included: 
\begin{itemize}
    \item Alpha-to-gamma power ratio
    \item Relative band power normalization
    \item Temporal co-modulation metrics
\end{itemize}

\subsection{Recurrence Plot Construction and Nonlinear Representation}
Given a band-limited EEG time series $x(t)$, phase-space reconstruction forms state vectors
\begin{equation}
\mathbf{x}_i = [x(i),\, x(i+\tau),\,\dots,\,x(i+(m-1)\tau)],
\label{eq:embed}
\end{equation}
where $m$ is the embedding dimension, $\tau$ is the delay, and $i=1,\dots,N$ indexes reconstructed states. The recurrence matrix is
\begin{equation}
R_{i,j}=
\begin{cases}
1,& \|\mathbf{x}_i-\mathbf{x}_j\|_2 \le \epsilon,\\
0,& \text{otherwise},
\end{cases}
\label{eq:rp}
\end{equation}
where $\epsilon$ is the recurrence threshold, RPs were computed separately for alpha and gamma signals and resized to fixed-resolution grayscale images for CNN input. This representation encodes nonlinear recurrence structure (e.g., periodicity and regime transitions) that complements time-frequency features.

In the context of Fragile X Syndrome, abnormal recurrence structures, such as increased recurrence density, disrupted diagonal line structure, and irregular texture, are hypothesized to reflect cortical hyperexcitability, reduced inhibitory control, and altered oscillatory regulation, particularly in the gamma band. By transforming EEG time series into recurrence plot images, this approach enables deep learning models to learn nonlinear dynamical features without explicit hand-crafted descriptors, providing a robust and interpretable representation for detecting EEG abnormalities associated with FXS.

\subsection{CNN-Based Spatial–Spectral and Nonlinear Feature Learning}
Convolutional Neural Networks (CNNs) were employed to learn discriminative spatial, spectral, and nonlinear features from structured EEG representations derived from alpha and gamma band signals. CNNs are particularly well suited for this task due to their ability to exploit local spatial correlations and hierarchical feature composition in two-dimensional inputs such as spectrograms and recurrence plots.
\subsubsection{Input Representations}
CNN inputs consisted of two complementary representations:
\begin{itemize}
    \item Time–frequency representations, obtained by computing short-time Fourier transform (STFT)–based spectrograms from band-limited EEG signals.
    \item Recurrence plot images, constructed independently for alpha and gamma band signals as described in Section 3.5.
\end{itemize}
Each input was resized to a fixed spatial resolution and normalized to ensure numerical stability during training. When multiple channels were used, representations were either concatenated along the channel dimension or aggregated using region-wise averaging to reduce dimensionality.

\subsubsection{CNN Architecture}
The CNN architecture comprised a sequence of convolutional blocks, each consisting of:
\begin{itemize}
    \item A convolutional layer with small receptive fields
    \item Rectified Linear Unit (ReLU) activation
    \item Batch normalization to stabilize training
    \item Max pooling to reduce spatial resolution and improve translational invariance.
\end{itemize}
Flattening layers were applied following the final convolutional block, and fully connected layers were used to project features into a compact latent space.

\subsubsection{Learning Nonlinear EEG Dynamics}
Recurrence plots encode nonlinear state recurrence patterns as spatial textures, enabling CNNs to learn dynamical features such as periodicity, laminar states, and transitions between oscillatory regimes. By jointly learning from spectrograms and recurrence plots, CNNs capture both frequency-domain energy distributions and nonlinear dynamical signatures, providing a richer representation than conventional spectral features alone.

\subsubsection{CNN-Based Classification Objective}
The learned feature embedding $\mathbf{Z}_{\mathrm{CNN}}$ was passed to a fully connected classification layer:

\begin{equation}
    \hat{y} = \sigma\left(\mathbf{W}_{\mathrm{CNN}} \mathbf{Z}_{\mathrm{CNN}} + \mathbf{b}_{\mathrm{CNN}}\right)
\end{equation}
where $\hat{y}$ denotes the predicted class probability, $\mathbf{W}_{\mathrm{CNN}}$ and $\mathbf{b}_{\mathrm{CNN}}$ are learnable parameters, and  $\sigma$ is the sigmoid or softmax activation function. The CNN was trained using cross-entropy loss and optimized via stochastic gradient descent with adaptive learning rates.

\subsubsection{Role of CNNs in the Overall Framework}
Within the proposed framework, CNNs serve as the primary mechanism for learning spatial–spectral and nonlinear EEG features that characterize alpha–gamma abnormalities in Fragile X Syndrome. These features are used directly for classification or passed to subsequent temporal modeling layers in hybrid CNN–LSTM architectures, enabling end-to-end learning of complex EEG patterns.

\subsubsection{LSTM-Based Temporal Modeling of EEG Dynamics}
While CNNs capture spatial–spectral and nonlinear structure from static representations, Long Short-Term Memory (LSTM) networks were employed to explicitly model the temporal evolution of EEG oscillations, which is critical for detecting abnormal temporal modulation and instability in Fragile X Syndrome.

\subsubsection{Temporal Feature Sequences}
For each EEG recording, temporal feature sequences were constructed from alpha and gamma band signals, including band-limited power, peak frequency, and cross-band ratios, segmented into fixed-duration windows. Let:
\begin{equation}
    \mathbf{S} = \{ s_t \}_{t=1}^{T'}
\end{equation}
where $\mathbf{S}$ denotes the temporal feature sequence,$s_t$ is the feature vector at time step 
$t$, and $T'$ is the number of temporal windows.

\subsubsection{LSTM Architecture}
The LSTM network processes the temporal sequence $\mathbf{S}$ to learn long-range dependencies and temporal patterns. The LSTM hidden state update is given by:
\begin{equation}
    \mathbf{h}_t = \mathcal{F}_{\mathrm{LSTM}}(s_t, \mathbf{h}_{t-1})
\end{equation}
where $\mathbf{h}_t$ is the hidden state at time $t$, $\mathbf{h}_{t-1}$ is the previous hidden state, and $ \mathcal{F}_{\mathrm{LSTM}}$ denotes the LSTM cell update function. The final LSTM embedding is defined as:
\begin{equation}
    \mathbf{Z}_{\mathrm{LSTM}} = \mathbf{h}_{T'}
\end{equation}
where $\mathbf{Z}_{\mathrm{LSTM}}$ is the learned temporal representation, and $T'$ is is the total number of temporal steps.

\subsection{Hybrid CNN–LSTM Architecture}
To jointly model spatial–spectral, nonlinear, and temporal EEG characteristics, a hybrid CNN–LSTM architecture was employed. This architecture integrates CNN-based feature extraction from recurrence plots or spectrograms with LSTM-based temporal modeling.

\subsubsection{3.8.1 Feature Fusion Strategy}
Let $\mathbf{R}$ denote a recurrence plot representation and $\mathbf{S}$ the corresponding temporal feature sequence. CNN-based feature extraction is first applied:

\begin{equation}
    \mathbf{Z}_{\mathrm{CNN}} = \mathcal{F}_{\mathrm{CNN}}(\mathbf{R})
\end{equation}
where $\mathbf{R}$ denotes the recurrence plot,  and $\mathbf{Z}_{\mathrm{CNN}}$is the CNN feature embedding. The hybrid representation is mapped to the output layer as:

\begin{equation}
    \hat{y} = \sigma(\mathbf{W}\mathbf{Z} + \mathbf{b})
\end{equation}
where $\hat{y}$ is the predicted class probability, $\mathbf{W}$ is the learnable weight matrix,  $\mathbf{b}$is the bias vector, and $\sigma$is the sigmoid or softmax activation. This end-to-end architecture enables simultaneous learning of nonlinear recurrence structure, spatial–spectral features, and temporal dependencies, providing a comprehensive representation of EEG abnormalities in Fragile X Syndrome.

\subsection{Model Training, Validation, and Evaluation Protocol}
\subsubsection{Training Procedure}
All models were trained using supervised learning with cross-entropy loss. Optimization was performed using the Adam optimizer with adaptive learning rates. Early stopping was employed based on validation loss to prevent overfitting.

\subsubsection{Subject-Independent Validation}
To ensure generalizability and avoid data leakage, subject-independent cross-validation was used. EEG segments from the same subject were restricted to a single data split (training, validation, or testing). Additionally, the model performance was evaluated using classification accuracy,
Precision, recall, and F1-score, and area under the receiver operating characteristic curve (AUC).
All models were implemented using modern deep learning frameworks and trained on GPU-enabled hardware. Hyperparameters were selected using validation-based tuning and held constant across comparative experiments to ensure fair evaluation.
\subsection{Model Implementation and Training Details}
All CNN, LSTM, and hybrid CNN-LSTM models were implemented using a modern deep learning framework and trained under identical experimental conditions to ensure fair comparison. Binary cross-entropy was used as the classification loss function for all architectures. Optimization was performed using the Adam optimizer with an initial learning rate of $10^{-3}$, which was adaptively reduced based on validation performance. Mini-batch training was employed, and models were trained for a fixed maximum number of epochs with early stopping triggered by validation loss convergence.

Regularization strategies included batch normalization in convolutional layers and dropout in fully connected layers to mitigate overfitting. Hyperparameters were selected using the validation set and then held constant across all comparative experiments. To ensure reproducibility and prevent information leakage, subject-independent data partitioning was strictly enforced, and all model evaluations were performed exclusively on held-out test subjects. Training and inference were executed on GPU-enabled hardware to ensure computational efficiency.

\section{Results}
\subsection{Quantitative Performance Comparison}
The hybrid CNN–LSTM architecture consistently outperformed single-modality models across all metrics, indicating that joint modeling of spatial–spectral, nonlinear, and temporal EEG features provides complementary information for identifying FXS-related abnormalities, shown in Table\ref{table1}.

\begin{table}[t]
\caption{Classification performance comparison across modeling approaches}
\label{table1}
\centering
\begin{tabular}{lccccc}
\hline
\textbf{Model} & \textbf{Accuracy (\%)} & \textbf{Precision (\%)} & \textbf{Recall (\%)} & \textbf{F1-score (\%)} & \textbf{AUC} \\
\hline
CNN& 82.4 $\pm$ 2.1 & 81.7 $\pm$ 2.4 & 80.9 $\pm$ 2.6 & 81.3 $\pm$ 2.2 & 0.88 \\
LSTM& 79.6 $\pm$ 2.5 & 78.2 $\pm$ 2.8 & 79.1 $\pm$ 2.7 & 78.6 $\pm$ 2.5 & 0.85 \\
\textbf{Hybrid} & \textbf{86.9 $\pm$ 1.8} & \textbf{86.1 $\pm$ 2.0} & \textbf{85.4 $\pm$ 2.1} & \textbf{85.7 $\pm$ 1.9} & \textbf{0.92} \\
\hline
\end{tabular}
\end{table}

\subsection{Contribution of Alpha and Gamma Bands}
To assess the relative contribution of alpha and gamma oscillations, models were trained using individual frequency bands and their combination. Gamma-band features provided stronger discriminative power than alpha alone; however, the combination of alpha and gamma features yielded the highest performance, supporting the hypothesis that alpha–gamma dysregulation is a key EEG marker of FXS, shown in Table.\ref{table2}.

\begin{table}[t]
\caption{Impact of frequency band selection on hybrid CNN-LSTM model performance}
\label{table2}
\centering
\begin{tabular}{lccc}
\hline
\textbf{Frequency Band} & \textbf{Accuracy (\%)} & \textbf{F1-score (\%)} & \textbf{AUC} \\
\hline
Alpha only (8-12 Hz) & 78.3 & 77.9 & 0.84 \\
Gamma only (30-100 Hz) & 83.5 & 83.1 & 0.89 \\
\textbf{Alpha \& Gamma} & \textbf{86.9} & \textbf{85.7} & \textbf{0.92} \\
\hline
\end{tabular}
\end{table}

\subsection{Effect of Recurrence Plot Representation}
To evaluate the impact of nonlinear dynamical representations, CNN models were trained with and without recurrence plots. The inclusion of recurrence plots improved classification performance, indicating that nonlinear dynamical features provide complementary information beyond conventional time–frequency representations, as shown in the Table. \ref{table3}.

\begin{figure}[htbp]
\centerline{\includegraphics[width =0.85\linewidth]{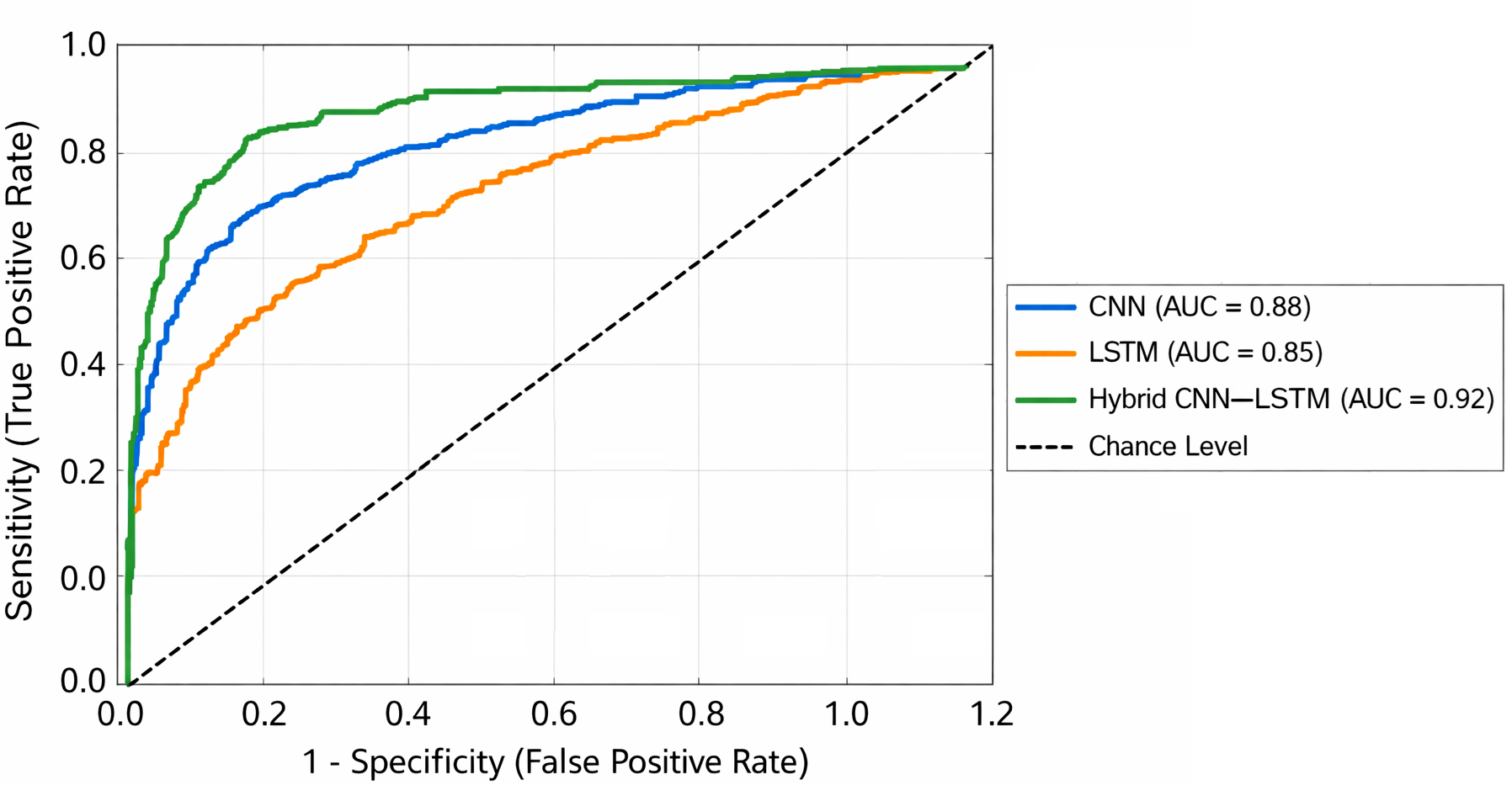}}
\caption{ROC curves for CNN, LSTM, and Hybrid CNN models}
\label{fig4}
\end{figure}

\begin{table}[t]
\caption{Effect of recurrence plot representation on CNN performance}
\label{table3}
\centering
\begin{tabular}{lccc}
\hline
\textbf{Representation} & \textbf{Accuracy (\%)} & \textbf{F1-score (\%)} & \textbf{AUC} \\
\hline
Spectrogram only & 79.8 & 79.2 & 0.85 \\
\textbf{Spectrogram \& RP} & \textbf{82.4} & \textbf{81.3} & \textbf{0.88} \\
\hline
\end{tabular}
\end{table}
Fig.\ref{fig4} illustrates the ROC curves for the three model variants. The hybrid CNN–LSTM model achieved the highest AUC, demonstrating superior discriminative capability across operating thresholds.

\subsection{Summary of Experimental Findings}
The experimental results demonstrate that the Hybrid CNN–LSTM architectures outperform single-modality models. Gamma-band abnormalities are particularly discriminative in FXS EEG; Alpha–gamma integration improves classification accuracy; and Recurrence plots enhance model performance by capturing nonlinear EEG dynamics.
These findings support the effectiveness of the proposed framework for modeling EEG abnormalities associated with Fragile X Syndrome and motivate its application in translational and clinical research contexts.

\section{Discussion and Neurophysiological Interpretation}
The experimental results demonstrate that EEG abnormalities in Fragile X Syndrome (FXS) are best characterized by a multi-scale modeling approach that integrates spatial–spectral, temporal, and nonlinear dynamical information. The hybrid CNN–LSTM architecture consistently outperformed single-modality models, indicating that FXS-related EEG alterations are not confined to isolated frequency bands or static features. In particular, gamma-band activity emerged as the most discriminative component, consistent with prior evidence of cortical hyperexcitability and impaired excitation–inhibition balance in FXS. While alpha-band features alone were less predictive, their integration with gamma dynamics significantly improved performance, supporting the hypothesis that alpha–gamma dysregulation reflects disrupted inhibitory control and large-scale network coordination in FXS.

The inclusion of recurrence plot representations further enhanced classification performance, highlighting the importance of nonlinear EEG dynamics in FXS. These representations capture the stability of the oscillatory regime and state transitions that are not accessible through linear spectral analysis alone. From a neurophysiological perspective, abnormal recurrence structures may reflect unstable cortical oscillatory states and reduced inhibitory gating, particularly within gamma-generating circuits. Collectively, these findings support the use of deep learning frameworks that combine temporal modeling and nonlinear representations for studying neurodevelopmental disorders and suggest a promising pathway toward automated, scalable EEG biomarkers for translational and clinical research in Fragile X Syndrome.

\subsection {Clinical Impact Statement}
This framework enables automated, objective characterization of EEG abnormalities in Fragile X Syndrome, supporting scalable biomarker development for diagnosis, patient stratification, and treatment monitoring in clinical and translational research settings.

\section{Conclusion and Future Work}
This paper presented a multi-representation deep learning framework for analyzing EEG abnormalities in Fragile X Syndrome, integrating convolutional and recurrent neural networks with nonlinear recurrence plot analysis. By explicitly modeling alpha and gamma oscillatory activity, the proposed approach demonstrated that joint spatial–spectral, temporal, and nonlinear representations provide superior discrimination between individuals with FXS and typically developing controls compared to single-modality models. The results support the hypothesis that FXS-related EEG abnormalities are multi-scale in nature and highlight the value of deep learning methods for capturing complex neurophysiological patterns beyond conventional spectral analysis.

Future work will focus on expanding the framework to larger and more diverse cohorts, incorporating connectivity and cross-frequency coupling measures, and improving interpretability through explainable AI techniques. Longitudinal studies and treatment-response analyses will also be explored to evaluate the potential of the proposed approach as a robust translational EEG biomarker for Fragile X Syndrome and related neurodevelopmental disorders.
%
%
%
%
\bibliographystyle{ieeetr}
\bibliography{references}

@ARTICLE{Cornish2004-vr,
  title    = "Annotation: Deconstructing the attention deficit in fragile {X}
              syndrome: a developmental neuropsychological approach",
  author   = "Cornish, K M and Turk, J and Wilding, J and Sudhalter, V and
              Munir, F and Kooy, F and Hagerman, R", 
  journal  = "J Child Psychol Psychiatry",
  volume   =  45,
  number   =  6,
  pages    = "1042--1053",
  month    =  sep,
  year     =  2004,
  address  = "England",
  language = "en"
}

@ARTICLE{Richter2021-uo,
  title    = "The molecular biology of {FMRP}: new insights into fragile {X}
              syndrome",
  author   = "Richter, Joel D and Zhao, Xinyu",
  journal  = "Nature Reviews Neuroscience",
  volume   =  22,
  number   =  4,
  pages    = "209--222",
  month    =  apr,
  year     =  2021
}

@article{moskowitz2015uncovering,
  title={Uncovering the evidence for behavioral interventions with individuals with fragile X syndrome: A systematic review},
  author={Moskowitz, Lauren J and Jones, Emily A},
  journal={Research in developmental disabilities},
  volume={38},
  pages={223--241},
  year={2015},
  publisher={Elsevier}
}

@article{knyazev2012eeg,
  title={EEG correlates of spontaneous self-referential thoughts: a cross-cultural study},
  author={Knyazev, Gennady G and Savostyanov, Alexander N and Volf, Nina V and Liou, Michelle and Bocharov, Andrey V},
  journal={International Journal of Psychophysiology},
  volume={86},
  number={2},
  pages={173--181},
  year={2012},
  publisher={Elsevier}
}

@ARTICLE{Wilkinson2021-pf,
  title    = "Increased aperiodic gamma power in young boys with Fragile {X}
              Syndrome is associated with better language ability",
  author   = "Wilkinson, Carol L and Nelson, Charles A",
  journal  = "Mol Autism",
  volume   =  12,
  number   =  1,
  pages    = "17",
  month    =  feb,
  year     =  2021,
  address  = "England",
  language = "en"
}

@ARTICLE{Liang2022-rn,
  title    = "Abnormal Brain Oscillations in Developmental Disorders:
              Application of Resting State {EEG} and {MEG} in Autism Spectrum
              Disorder and Fragile {X} Syndrome",
  author   = "Liang, Sophia and Mody, Maria",
  journal  = "Front Neuroimaging",
  volume   =  1,
  pages    = "903191",
  month    =  may,
  year     =  2022,
  address  = "Switzerland",
  language = "en"
}

@ARTICLE{Castren2003-uh,
  title    = "Augmentation of auditory {N1} in children with fragile {X}
              syndrome",
  author   = "Castr{\'e}n, Maija and P{\"a}{\"a}kk{\"o}nen, Ari and Tarkka, Ina
              M and Ryyn{\"a}nen, Markku and Partanen, Juhani",  
  journal  = "Brain Topogr",
  volume   =  15,
  number   =  3,
  pages    = "165--171",
  year     =  2003,
  address  = "United States",
  language = "en"
}

@Article{Knoth2018,
author={Knoth, Inga Sophia
and Lajnef, Tarek
and Rigoulot, Simon
and Lacourse, Karine
and Vannasing, Phetsamone
and Michaud, Jacques L.
and Jacquemont, S{\'e}bastien
and Major, Philippe
and Jerbi, Karim
and Lipp{\'e}, Sarah},
title={Auditory repetition suppression alterations in relation to cognitive functioning in fragile X syndrome: a combined EEG and machine learning approach},
journal={Journal of Neurodevelopmental Disorders},
year={2018},
month={Jan},
day={29},
volume={10},
number={1},
pages={4},
issn={1866-1955},
doi={10.1186/s11689-018-9223-3},
url={https://doi.org/10.1186/s11689-018-9223-3}
}

@Article{Ethridge2024,
author={Ethridge, Lauren E.
and Pedapati, Ernest V.
and Schmitt, Lauren M.
and Norris, Jordan E.
and Auger, Emma
and De Stefano, Lisa A.
and Sweeney, John A.
and Erickson, Craig A.},
title={Validating brain activity measures as reliable indicators of individual diagnostic group and genetically mediated sub-group membership in Fragile X Syndrome},
journal={Scientific Reports},
year={2024},
month={Oct},
day={03},
volume={14},
number={1},
pages={22982},
issn={2045-2322},
doi={10.1038/s41598-024-72935-6},
url={https://doi.org/10.1038/s41598-024-72935-6}
}

@ARTICLE{Norris2025-vc,
  title    = "{ROC} Analysis of Biomarker Combinations in Fragile {X}
              {Syndrome-Specific} Clinical Trials: Evaluating Treatment
              Efficacy via Exploratory Biomarkers",
  author   = "Norris, Jordan E and Berry-Kravis, Elizabeth M and Harnett, Mark
              D and Reines, Scott A and Smith, Melody R and Auger, Emma K and
              Outterson, Abigail H and Furman, Jeremiah and Gurney, Mark E and
              Ethridge, Lauren E",
  journal  = "Transl Psychiatry",
  volume   =  15,
  number   =  1,
  pages    = "323",
  month    =  aug,
  year     =  2025,
  address  = "United States",
  language = "en"
}

@ARTICLE{Ethridge2017-si,
  title     = "Neural synchronization deficits linked to cortical
               hyper-excitability and auditory hypersensitivity in fragile {X}
               syndrome",
  author    = "Ethridge, Lauren E and White, Stormi P and Mosconi, Matthew W
               and Wang, Jun and Pedapati, Ernest V and Erickson, Craig A and
               Byerly, Matthew J and Sweeney, John A",
  journal   = "Mol. Autism",
  publisher = "Springer Science and Business Media LLC",
  volume    =  8,
  number    =  1,
  month     =  dec,
  year      =  2017,
  copyright = "http://creativecommons.org/licenses/by/4.0/",
  language  = "en"
}

@article{van2012auditory,
  title={Auditory change detection in fragile X syndrome males: a brain potential study},
  author={Van der Molen, MJW and Van der Molen, MW and Ridderinkhof, KR and Hamel, BCJ and Curfs, LMG and Ramakers, GJA},
  journal={Clinical neurophysiology},
  volume={123},
  number={7},
  pages={1309--1318},
  year={2012},
  publisher={Elsevier}
}

@article{Kenny2022Biomarker,
  author  = {Kenny, Alex and Ethridge, L. Erin and Pedapati, Ernest V. and Erickson, Craig A.},
  title   = {Electroencephalography as a Translational Biomarker and Outcome Measure in Fragile X Syndrome},
  journal = {Frontiers in Integrative Neuroscience},
  volume  = {16},
  pages   = {905386},
  year    = {2022},
  doi     = {10.3389/fnint.2022.905386}
}

@article{Ethridge2019Auditory,
  author  = {Ethridge, L. Erin and Pedapati, Ernest V. and Takarae, Yukari and Mosconi, Matthew W. and Erickson, Craig A.},
  title   = {Auditory EEG Biomarkers in Fragile X Syndrome: Clinical Relevance and Neural Mechanisms},
  journal = {Frontiers in Integrative Neuroscience},
  volume  = {13},
  pages   = {41},
  year    = {2019},
  doi     = {10.3389/fnint.2019.00041}
}

@Article{Wang2017,
author={Wang, Jun
and Ethridge, Lauren E.
and Mosconi, Matthew W.
and White, Stormi P.
and Binder, Devin K.
and Pedapati, Ernest V.
and Erickson, Craig A.
and Byerly, Matthew J.
and Sweeney, John A.},
title={A resting EEG study of neocortical hyperexcitability and altered functional connectivity in fragile X syndrome},
journal={Journal of Neurodevelopmental Disorders},
year={2017},
month={Mar},
day={14},
volume={9},
number={1},
pages={11},
issn={1866-1955},
doi={10.1186/s11689-017-9191-z},
url={https://doi.org/10.1186/s11689-017-9191-z}
}

@article{LIU2023100070,
title = {Reliability of resting-state electrophysiology in fragile X syndrome},
journal = {Biomarkers in Neuropsychiatry},
volume = {9},
pages = {100070},
year = {2023},
issn = {2666-1446},
doi = {https://doi.org/10.1016/j.bionps.2023.100070},
url = {https://www.sciencedirect.com/science/article/pii/S2666144623000102},
author = {Rui Liu and Ernest V. Pedapati and Lauren M. Schmitt and Rebecca C. Shaffer and Elizabeth G. Smith and Kelli C. Dominick and Lisa A. DeStefano and Grace Westerkamp and Paul Horn and John A. Sweeney and Craig A. Erickson}
}

@article{pedapati2025frontal,
  title={Frontal cortex hyperactivation and gamma desynchrony in Fragile X syndrome: Correlates of auditory hypersensitivity},
  author={Pedapati, Ernest V and Ethridge, Lauren E and Liu, Yanchen and Liu, Rui and Sweeney, John A and DeStefano, Lisa A and Miyakoshi, Makoto and Razak, Khaleel and Schmitt, Lauren M and Moore, David R and others},
  journal={PLoS One},
  volume={20},
  number={5},
  pages={e0306157},
  year={2025},
  publisher={Public Library of Science San Francisco, CA USA}
}
\end{document}